\documentclass[11pt,a4paper]{article}
\usepackage[hyperref]{acl2018}
\usepackage{times}
\usepackage{latexsym}
\usepackage{graphicx}
\usepackage{comment,subcaption}
\usepackage[font={small}]{caption}

\usepackage{pgfplots}
\usepackage{tikz} 

\usepackage{url}

\aclfinalcopy % Uncomment this line for the final submission

\usepackage{times}
\usepackage{latexsym}

\usepackage{url}

\usepackage{xcolor}

\definecolor{correctingcolor}{RGB}{0,109,44}
\definecolor{skepticalcolor}{RGB}{116,196,118}
\definecolor{trustingcolor}{RGB}{186,228,179}
\definecolor{questioningcolor}{RGB}{0,90,50}

\definecolor{lineplotscolor}{RGB}{0,109,44}

\definecolor{D}{RGB}{0,0,0} 
\definecolor{V}{RGB}{4,90,141}%3,72,23} 
\colorlet{CB}{blue!75!black} 
\colorlet{CS}{orange!80!black} 
\colorlet{P}{black}%{red!75!black} 

\colorlet{answer_color}{blue!50!white} 
\colorlet{appreciation_color}{blue!90!white} 
\colorlet{elaborate_color}{blue!75!black} 
\colorlet{humor_color}{blue!40!black} 
\colorlet{question_color}{black} 
\colorlet{no_label_color}{gray}

\definecolor{T_Reddit}{RGB}{102,194,164} 
\definecolor{DID_Reddit}{RGB}{44,162,95} 
\definecolor{DED_Reddit}{RGB}{102,194,164} 

\definecolor{T_Twitter}{RGB}{102,194,164} 
\definecolor{DID_Twitter}{RGB}{44,162,95} 
\definecolor{DED_Twitter}{RGB}{102,194,164}

\usepackage{tikz} 
\usepackage{pgfplots}

\newcommand*{\ie}{{\em i.e.,~}}
\newcommand*{\eg}{{\em e.g.,~}}
\newcommand*{\etal}{et al.~}
\newcommand*{\ignore}[1]{}

\title{Identifying and Understanding User Reactions to \\Deceptive and Trusted Social News Sources}

\author{Maria Glenski~~~~~~~~~~~~~~Tim Weninger \\
  Computer Science and Engineering \\
  University of Notre Dame \\
  Notre Dame, IN 46556 \\ 
  {\tt \{mglenski, tweninge\}@nd.edu} \\\And 
  Svitlana Volkova \\
  Data Sciences and Analytics Group \\
  Pacific Northwest National Laboratory \\ 
  Richland, WA 99354 \\
  {\tt svitlana.volkova@pnnl.gov} \\}

\date{}

\begin{document}
\maketitle
\begin{abstract}
  In the age of social news, it is important to understand the types of reactions that are evoked from news sources with various levels of credibility. In the present work we seek to better understand how users react to trusted and deceptive news sources across two popular, and very different, social media platforms. To that end, (1) we develop a model to classify user reactions into one of nine types, such as answer, elaboration, and question, etc, and (2) we measure the speed and the type of reaction for trusted and deceptive news sources for 10.8M Twitter posts and 6.2M Reddit comments. We show that there are significant differences in the speed and the type of reactions between trusted and deceptive news sources on Twitter, but far smaller differences on Reddit. 
\end{abstract}

\section{Introduction}

As the reliance on social media as a source of news increases and the reliability of sources is increasingly debated, it is important to understand how users react to various sources of news. Most studies that investigate misinformation spread in social media focus on individual events and the role of the network structure in the spread~\cite{qazvinian2011rumor,wu2015false,kwon2017rumor} or detection of false information~\cite{rath2017retweet}. 
These studies have found that the size and shape of misinformation cascades within a social network depends heavily on the initial reactions of the users. Other work has focused on the language of misinformation in social media~\cite{rubin2016fake,rashkin2017truth,mitra2017parsimonious,wang2017liar,karadzhov2017we,volkova2017separating} to detect types of deceptive news. 

As an alternative to studying newsworthy events one at a time~\cite{starbird2017examining}, the current work applies linguistically-infused models to predict user reactions to deceptive and trusted news {\em sources}. Our analysis reveals differences in reaction types and speed across two social media platforms --- Twitter and Reddit.

The first metric we report is the reaction type. Recent studies have found that 59\% of bitly-URLs on Twitter are shared without ever being read~\cite{gabielkov2016social}, and 73\% of Reddit posts were voted on without reading the linked article~\cite{glenski2017consumers}. Instead, users tend to rely on the commentary added to retweets or the comments section of Reddit-posts for information on the content and its credibility. Faced with this reality, we ask: what kind of reactions do users find when they browse sources of varying credibility? 
Discourse acts, or speech acts, can be used to identify the \textit{use} of language within a conversation, \eg agreement, question, or answer. Recent work by Zhang~\etal\shortcite{zhang2017characterizing} classified Reddit comments by their primary discourse act (\eg question, agreement, humor), and further analyzed patterns from these discussions. 

The second metric we report is reaction speed. A study by Jin \etal \shortcite{jin2013epidemiological} found that trusted news {\em stories} spread faster than misinformation or rumor; Zeng \etal\shortcite{zeng2016rumors} found that tweets which deny rumors had shorter delays than tweets of support. Our second goal is to determine if these trends are maintained for various types of news sources on Twitter and Reddit.  

Hence, the contributions of this work are two-fold: (1) we develop a linguistically-infused neural network model to classify reactions in social media  posts, and (2) we apply our model to label 10.8M Twitter posts and 6.2M Reddit comments in order to evaluate the speed and type of user reactions to various news sources.

\section{Reaction Type Classification} 
In this section, we describe our approach to classify user reactions into one of eight types of discourse: agreement, answer, appreciation, disagreement, elaboration, humor, negative reaction, or question, or as none of the given labels, which we call ``other'', using linguistically-infused neural network models.

\subsection{Reddit Data}
We use a manually annotated Reddit dataset from Zhang \etal\shortcite{zhang2017characterizing} to train our reaction classification model. Annotations from 25 crowd-workers labelled the primary discourse act for 101,525 comments within 9,131 comment threads on Reddit. The Reddit IDs, but not the text content of the comments themselves, were released with the annotations. So we collected the content of Reddit posts and comments from a public archive of Reddit posts and comments.\footnote{\url{bigquery.cloud.google.com/dataset/fh-bigquery:reddit_posts | reddit_comments}}  Some content was deleted prior to archival, so the dataset shown in Table~\ref{tab:annotation_data_breakdown} is a subset of the original content. Despite the inability to capture all of the original dataset, Table~\ref{tab:annotation_data_breakdown} shows a similar distribution between our dataset and the original.

	\begin{table}[!ht]
		\centering
		\small
		%\reduceSpaceAroundFigure
		\begin{tabular}{l|r@{\hskip6pt}r@{\hskip10pt}r@{\hskip6pt}r} 
    & \multicolumn{2}{r}{Zhang \etal}  & \multicolumn{2}{r}{Present work} \\ 
   Reaction Type &    \#  &    \% &   \# & \% \\
   \hline
            agreement  &     5,054    &    4.73 &  3,857  &  4.61   \\ 
               answer  &    41,281    &   38.63 & 32,561  & 38.94   \\
         appreciation  &     8,821    &    8.25 &  6,973  &  8.34   \\
         disagreement  &     3,430    &    3.21 &  2,654  &  3.17   \\
          elaboration  &    19,315    &   18.07 & 14,966  & 17.90   \\
                humor  &     2,358    &    2.21 &  1,878  &  2.25   \\
    negative reaction  &     1,901    &    1.78 &  1,473  &  1.76   \\
                other  &     1,979    &    1.85 &  1,538  &  1.84   \\
             question  &    10,568    &    9.89 &  8,194  &  9.80   \\
    no majority label  &    12,162    &   11.38 &  9532  & 11.40   \\
\hline
               \textbf{Total} &     \textbf{106,869} &     \textbf{100} &  \textbf{83,   626} &  \textbf{100}\\
		\end{tabular}
		\caption{Summary of the training data we recovered compared to the data collected by Zhang \etal\shortcite{zhang2017characterizing} reported as  distributions of comments across reaction types.}
		\label{tab:annotation_data_breakdown}
		\vspace{-0.3cm}
	\end{table}
 
\subsection{Model}

\begin{figure}[t!] 
 
 \begin{tikzpicture}[scale=0.42, transform shape]

%merged network
 \draw[->,ultra thick] (-2,4.5) -- (-2,5.2);
\node at (-2,5.53) {\LARGE Predicted Label};%
 
\draw  (-3.5,4.3) rectangle (-0.5,3.5); %activation layer
\node[right] at (-0.2,4.1) {\LARGE Probability Activation };%
\node[right] at (-0.2,3.6) {\LARGE Layer (soft max)};%

\draw  (-3,3) rectangle (-4,2.2); %left dense
\draw[loosely dotted, ultra thick] (-2.5,2.6) edge (-1.5,2.6);
\draw  (-1,3) rectangle (0,2.2); % right dense  
%\node[right] at (0.5,2.8) {\LARGE Dense Layer};%
%\node[right] at (0.5,2.3) {\LARGE (100 units)};%
\node[left] at (-4.2,2.8) {\LARGE Dense Layer};%
\node[left] at (-4.2,2.3) {\LARGE (100 units)};%

\draw (-3.5,3) edge (-2.5,3.5) {};
\draw (-0.5,3) edge (-1.5,3.5) {};

\draw  (-3.5,1.7) rectangle (-0.5,0.9);
\node[right] at (-0.2,1.3) {\LARGE Tensor Concatenation};%

\draw (-2.5,1.7) edge (-3.5,2.2) {};
\draw (-1.5,1.7) edge (-0.5,2.2) {};

%% Text subnetwork
%Max pooling 
\draw  (-1.5,0) rectangle (-0.5,-0.8) node (v1) {}; %left max pooling
\draw[loosely dotted, ultra thick] (0,-.4) edge (1,-.4);
\draw  (1.5,0) rectangle (2.5,-0.8); %right max pooling
\node[right] at (3,-0.1) {\LARGE Max Pooling Layer};%
\node[right] at (3,-0.7) {\LARGE (1D, Pool Size=3)};%

\draw (-1.5,0.9) edge (-1,0) {};
\draw (-1,0.9) edge (2,0) {};

%CNN 2
\draw  (-1.75,-1.5) rectangle (-0.75,-2.5); %CNN 2 left
\draw  (1.75,-1.5) rectangle (2.75,-2.5); %CNN 2 right
\node[right] at (3,-1.7) {\LARGE Convolutional };
\node[right] at (3,-2.25) {\LARGE Layer (100 units)};%

\draw (-1.25,-1.5) edge (-1,-0.8) {};
\draw (2.25,-1.5) edge (2,-0.8) {};
\draw  (-0.5,-0.8) edge (1.75,-1.5);
\draw  (-0.75,-1.5) edge (1.5,-0.8);
\draw[->,thick] (-0.4,-2) -- (0,-2);
\draw[loosely dotted, ultra thick] (0.1,-2) edge (0.9,-2);
\draw[->,thick] (0.9,-2) -- (1.4,-2);

%CNN 1
\draw  (-1.75,-3.5) rectangle (-0.75,-4.5); %CNN 1 left
\draw  (1.75,-3.5) rectangle (2.75,-4.5); %CNN 1 right 
\node[right] at (3,-3.7) {\LARGE Convolutional };
\node[right] at (3,-4.25) {\LARGE Layer (100 units)};%

\draw (-1.25,-3.5) edge (-1.25,-2.5) {};
\draw (2.25,-3.5) edge (2.25,-2.5) {};

\draw  (-1,-2.5) edge (2,-3.5);
\draw  (-1,-3.5) edge (2,-2.5);

\draw[->,thick] (-0.4,-4) -- (0,-4);
\draw[loosely dotted, ultra thick] (0.1,-4) edge (0.9,-4);
\draw[->,thick] (0.9,-4) -- (1.4,-4);

%embedding layer
\draw  (-1.5,-6) rectangle (-0.5,-5.2); %embedding left
\draw[loosely dotted, ultra thick] (0,-5.6) edge (1,-5.6);
\draw  (1.5,-6) rectangle (2.5,-5.2); %embedding right
\node[right] at (3,-5.35) {\LARGE Embedding Layer};%
\node[right] at (3,-5.85) {\LARGE (200 units)};% 

\draw (-1.25,-4.5) edge (-1,-5.2) {};
\draw (2.25,-4.5) edge (2,-5.2) {};
\draw  (-0.75,-4.5) edge (1.75,-5.2);
\draw  (-0.75,-5.2) edge (1.75,-4.5);

\draw[->,thick] (0.5,-6.4) -- (0.5,-6);
\node[below] at (3.75,-6.4) {\LARGE Word Sequences};%

\draw  (-0.5,-6.5) rectangle (0,-7);
\draw  (0,-6.5) rectangle (0.5,-7);
\draw  (0.5,-6.5) rectangle (1,-7);
\draw  (1,-6.5) rectangle (1.5,-7);

%%x  -5
%Vector subnetwork
%connect to after merge model
\draw (-6,-3.7) edge (-3,0.9) {};
\draw ((-3,-3.7) edge (-2.5,0.9) {}; 

%dense 1
\draw  (-6.5,-6) rectangle (-5.5,-5.2); %dense 1 left
\draw[loosely dotted, ultra thick] (-5,-5.6) edge (-4,-5.6);
\draw  (-2.5,-6) rectangle (-3.5,-5.2); %dense 1 right
\node[left] at (-6.75,-5.35) {\LARGE Dense Layer};%
\node[left] at (-6.75,-5.85) {\LARGE (100 units)};%

\draw (-6,-4.5) edge (-6,-5.2) {};
\draw (-3,-4.5) edge (-3,-5.2) {};
\draw  (-5.75,-4.5) edge (-3.25,-5.2);
\draw  (-5.75,-5.2) edge (-3.25,-4.5);

%dense 1
\draw  (-6.5,-4.5) rectangle (-5.5,-3.7); %dense 1 left
\draw[loosely dotted, ultra thick] (-5,-4.1) edge (-4,-4.1);
\draw  (-2.5,-4.5) rectangle (-3.5,-3.7); %dense 1 right
\node[left] at (-6.75,-3.85) {\LARGE Dense Layer};%
\node[left] at (-6.75,-4.35) {\LARGE (100 units)};%

\draw[->,thick] (-4.5,-6.4) -- (-4.5,-6);
\node[below] at (-7.75,-6.4) {\LARGE LIWC Features};%

\draw  (-5.5,-6.5) rectangle (-5,-7);
\draw  (-5,-6.5) rectangle (-4.5,-7);
\draw  (-4.5,-6.5) rectangle (-4,-7);
\draw  (-4,-6.5) rectangle (-3.5,-7);

\draw[gray,dashed]  (-10.25,-3) rectangle (-2.25,-7.5);
\node[below right] at (-10.35,-7.5) {\LARGE \bf Vector Sub-Network};

\draw[gray,dashed]  (-2,0.5) rectangle (8,-7.5);
\node[below right] at (-2.1,-7.5) {\LARGE \bf Text Sub-Network};
\end{tikzpicture}

 %\vspace{-.1in}
 \caption{Architecture of neural network model used to predict reaction types.}
 \label{fig:model_architecture}
 \vspace{-0.35cm}
 \end{figure}
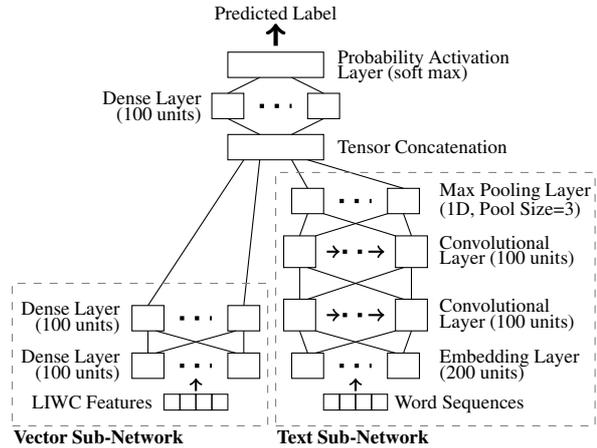

We develop a neural network architecture that relies on content and other linguistic signals extracted from reactions and parent posts, and takes advantage of a ``late fusion'' approach previously used effectively in vision tasks~\cite{karpathy2014large,park2016combining}. More specifically, we combine a text sequence sub-network with a vector representation sub-network as shown in Figure~\ref{fig:model_architecture}.  
 The text sequence sub-network consists of an embedding layer initialized with 200-dimensional GloVe embeddings~\cite{glove} followed by two 1-dimensional convolution layers, then a max-pooling layer followed by a dense layer. The vector representation sub-network consists of two dense layers. We incorporate information from both sub-networks through concatenated %\textbf{N}-dimensional 
 padded text sequences and vector representations of normalized Linguistic Inquiry and Word Count (LIWC) features~\cite{pennebaker2001linguistic} for the text of each post and its parent.

\subsection{Reaction Type Classification Results}
 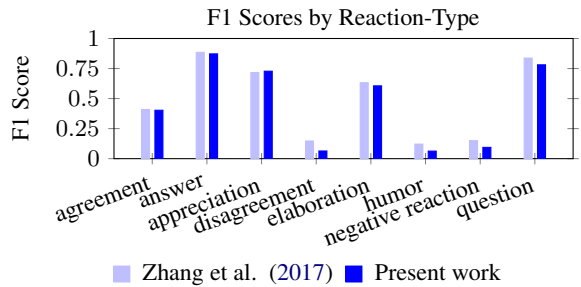
\begin{figure}[b!] 
  \vspace{-.2in}
  
  \small

\begin{tikzpicture}
	\begin{axis}[ 
	ybar,
	height=1.25in, width=3in,bar width=3pt,  
	legend style={at={(0.5,-1)},column sep =1.5mm,
		anchor=north,legend columns=-1,draw=none},
	ylabel={F1 Score}, 
	ymin=0,
	ymax=1, 
	symbolic x coords={ag,ans,ap,dis,el,hu,neg,q,o},
	xticklabels ={agreement,answer,appreciation,disagreement,elaboration,humor,negative reaction,question,other},
	xtick={ag,ans,ap,dis,el,hu,neg,q,o},tick align=center,
	xticklabel style = {rotate=20,anchor=north east},%,font=\tiny},
	title = {\small F1 Scores by Reaction-Type},
	title style={yshift=-1.5ex},
	xtick pos=left, 
	ytick= {0,0.25,0.5,0.75,1}, 
	]  
	 
	% zhang et al
	\addplot[blue!25!white, fill=blue!25!white ] coordinates { 
		(ag,	0.40899)%0.4	) 
        (ans,	0.885)%0.9	)
        (ap,	0.716999)%0.7	)
        (dis,	0.147)%0.1 )
        (el,	0.63199)%0.6	)
        (hu,	0.121)%0.1	)
        (neg,	0.15)%0.1	)
        (q,	0.83699)%0.8	) 

	}; 
	% us (best)
	\addplot[blue, fill=blue ] coordinates { 
		(ag,	0.403	) 
        (ans,	0.872	)
        (ap,	0.728	)
        (dis,	0.064 )
        (el,	0.606	)
        (hu,	0.063	)
        (neg,	0.094	)
        (q,	0.781	)

	};

	%\legend{Zhang \etal~\shortcite{zhang2017characterizing}, CNN}
	\end{axis}
\end{tikzpicture}

\vspace{-.1in}
\begin{tikzpicture} 
\begin{axis}[%
hide axis,
xmin=10,xmax=50,ymin=0,ymax=0.4,
legend style={draw=none,legend cell align=left,legend columns = 3, column sep = 1mm}
]
\addlegendimage{mark=square*,only marks,mark options={scale=1.5}, blue!25!white}
\addlegendentry{Zhang \etal~\shortcite{zhang2017characterizing}};
\addlegendimage{mark=square*,only marks,mark options={scale=1.5}, blue}
\addlegendentry{Present work};%CNN}; 
\end{axis}
\end{tikzpicture}
\vspace{-2.1in}
  
 \caption{Comparison of our model's performance, measured using F1 score, trained only on content features, with the performance reported by Zhang \etal\shortcite{zhang2017characterizing} trained on content, author, thread, structure, and community features.}
 \label{fig:model_performance}
 \end{figure}
 
As shown in Figure~\ref{fig:model_performance}, our linguistically-infused neural network model that relies solely on the content of the reaction and its parent has comparable performance to the more-complex CRF model by Zhang \etal\shortcite{zhang2017characterizing}, which relies on content as well as additional metadata like the author, thread (\eg the size of the the thread, the number of branches), structure (\eg the position within the thread), and community (\ie the subreddit in which the comment is posted).

\section{Measuring Reactions to Trusted and Deceptive News Sources}
%\vspace{-.5\baselineskip}
In this section, we present key results of our analysis of \textit{how often} and \textit{how quickly} users react to content from sources of varying credibility using the reaction types predicted by our linguistically-infused neural network model.

\subsection{Twitter and Reddit News Data} 
 
 We focus on \textit{trusted} news sources that provide factual information with no intent to deceive and \textit{deceptive} news sources. Deceptive sources are ranked by their intent to deceive as follows: clickbait (attention-grabbing, misleading, or vague headlines to attract an audience), conspiracy theory (uncorroborated or unreliable information to explain events or circumstances), propaganda (intentionally misleading information to advance a social or political agenda), and disinformation (fabricated or factually incorrect information meant to intentionally deceive readers). 

 Trusted, clickbait, conspiracy, and propaganda sources were previously compiled by Volkova et al.~\shortcite{volkova2017separating} through a combination of crowd-sourcing and public resources. Trusted news sources with Twitter-verified accounts were manually labeled and clickbait, conspiracy, and propaganda news sources were collected from several public resources that annotate suspicious news accounts\footnote{Example resources used by Volkova et al~\shortcite{volkova2017separating} to compile deceptive news sources: \url{http://www.fakenewswatch.com/},  \url{http://www.propornot.com/p/the-list.html} and others.}. We collected news sources identified as spreading disinformation by the European Union's East Strategic Communications Task Force from euvsdisinfo.eu. In total, there were 467 news sources: 251 trusted and 216 deceptive.

\begin{figure}[t!]
	\centering 
	\small    
	\small

\begin{tikzpicture}
	\begin{axis}[ 
	ybar,
	height=1.2in, width=3in,bar width=8pt,  
	legend style={at={(0.5,-0.2)},
		anchor=north,legend columns=-1,draw=none},
	ylabel={\# sources}, 
	ylabel style={yshift=-0.1in},  
	symbolic x coords={cb,cs,p,d},
	xticklabels ={ , , , }, 
	xtick={cb,cs,p,d},tick align=center,
	xticklabel style = {rotate=20,anchor=north east},%,font=\tiny}, 
	xtick pos=left, 
	%ymode = log, log origin y=infty, 
	]  
	 
	% reddit
	\addplot[blue, fill=blue] coordinates {  
        (cb, 19)
        (cs, 34) 
        (p, 75) 
        (d, 51) 
	}; 
	% twitter
	\addplot[blue!25!white, fill=blue!25!white] coordinates {  
	%\addplot[blue, fill=white, dashed, thick] coordinates {  
        (cb, 11)
        (cs, 13) 
        (p, 26) 
        (d, 50) 
	};  
  
	\end{axis}
\end{tikzpicture}
\vspace{-.05in}

\begin{tikzpicture}
	\begin{axis}[ 
	ybar,
	height=1.2in, width=3in,bar width=8pt,  
	legend style={at={(0.5,-0.5)},
		anchor=north,legend columns=-1,draw=none},
	ylabel={\# reactions}, 
	ylabel style={yshift=-0.1in}, 
	symbolic x coords={cb,cs,p,d},
	xticklabels ={Clickbait,Conspiracy,Propaganda,Disinformation}, 
	xtick={cb,cs,p,d},tick align=center,
	xticklabel style = {rotate=20,anchor=north east},%,font=\tiny}, 
	xtick pos=left, 
	ymode = log, log origin y=infty, 
	]

	% reddit
	\addplot[blue, fill=blue] coordinates {  
        (cb, 124392)
        (cs, 71053) 
        (p, 469177) 
        (d, 130908) 
	}; 
	% twitter
	\addplot[blue!25!white, fill=blue!25!white] coordinates {  
	%\addplot[blue, fill=white, dashed, thick] coordinates {  
        (cb, 40347)
        (cs, 126246) 
        (p, 609251) 
        (d, 3487732) 
	};  
	
	\end{axis}
\end{tikzpicture}

\vspace{-.1in}

\begin{tikzpicture} 
\begin{axis}[%
hide axis,
xmin=10,xmax=50,ymin=0,ymax=0.4,
legend style={draw=none,legend cell align=left,legend columns = -1, column sep = 1mm}%1mm}
]
\addlegendimage{mark=square*,only marks,mark options={scale=1.5}, blue, fill=blue}
\addlegendentry{Reddit Dataset};
\addlegendimage{mark=square*,only marks,mark options={scale=1.5}, blue!25!white, fill=blue!25!white}
%\addlegendimage{mark=square*,only marks,mark options={scale=1.5}, blue, fill=white, dashed, thick}
\addlegendentry{Twitter Dataset}; 
\addlegendimage{empty legend};
\addlegendentry{~~~~}; 
\end{axis}
\end{tikzpicture}

    \vspace{-2.1in}
	\caption{Distributions of Deceptive news sources and reactions to those sources (Reddit comments or tweets, respectively) for the Reddit and Twitter datasets across the four sub-categories of deceptive news sources.}
	\label{fig:deceptive_subcategories}
	\vspace{-0.3cm}
\end{figure}
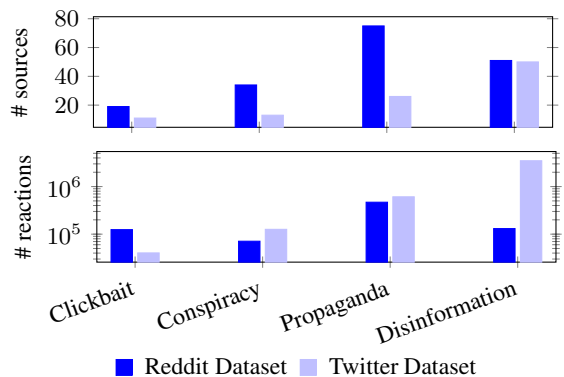

 We collected reaction data for two popular platforms, Reddit and Twitter, using public APIs over the 13 month period from January 2016 through January 2017. For our Reddit dataset, we collected all Reddit posts submitted during the 13 month period that linked to domains associated with one of our labelled news sources. Then we collected all comments that directly responded to those posts. For our Twitter dataset,  we collected all tweets posted in the 13 month period that explicitly @mentioned or directly retweeted content from a source and then assigned a label to each tweet based on the class of the source @mentioned or retweeted. A breakdown of each dataset by source type is shown in Table~\ref{tab:breakdown}. Figure~\ref{fig:deceptive_subcategories} illustrates the distribution of deceptive news sources and reactions across the four sub-categories of deceptive news sources.
  In our analysis, we consider the set of all deceptive sources and the set excluding the most extreme (disinformation).

	\begin{table}[t]
		\centering
		\small 
		\begin{tabular}{lrrr@{\hskip-6pt}}  
		    \multicolumn{3}{c}{\textit{Reddit Dataset}}\\
			Type              &\# Sources  & \# Comments \\ %& \# Posts
			\hline 
			%including disinfo 
            Trusted & 169 & 5,429,694 \\%& 828,863  
            Deceptive (no disinfo)  & 128 & 664,670 \\%& 135,294 
            Deceptive & 179 & 795,591 \\%& 155,575  
			%~~~~~~\textit{- Clickbait}         &       \textit{19}   &     \textit{124,392}  \\
			%~~~~~~\textit{- Conspiracy}        &       \textit{34}   &      \textit{71,053}  \\
			%~~~~~~\textit{- Propaganda}        &       \textit{75}   &     \textit{469,177}  \\
			%~~~~~~\textit{- Disinformation}    &       \textit{51}   &     \textit{130,908}  \\
            \hline
            %\bf Total (no disinformation)& \bf 297  & \bf 6,094,364 \\%& \bf 964,157
            \bf Total & \bf 348  & \bf 6,225,285 \\%& \bf 984,438
             
		    \multicolumn{3}{c}{}\\
		    \multicolumn{3}{c}{\textit{Twitter Dataset}}\\
			Type              &\# Sources & \# Tweets \\
			\hline 
			%including disinfo 
            Trusted  & 182 & 6,567,002  &       \\ 
            Deceptive (no disinfo)& 100 & 775,844  \\
            Deceptive & 150 & 4,263,576  &    \\
			%~~~~~~\textit{- Clickbait  }    &       \textit{11}  &     \textit{40,347}  \\
			%~~~~~~\textit{- Conspiracy }    &       \textit{13}  &    \textit{126,246}  \\
			%~~~~~~\textit{- Propaganda }    &       \textit{26}  &    \textit{609,251}  \\
			%~~~~~~\textit{- Disinformation } &      \textit{50}  &  \textit{3,487,732}  \\
            \hline
            %\bf Total (no disinformation)& \bf 282 & \bf 7,342,846 \\
            \bf Total & \bf 232 & \bf 10,830,578  & \\
             
		\end{tabular}
		
		\caption{Summary of Twitter and Reddit datasets used to measure the speed and types of reactions to Trusted and Deceptive news sources excluding (no disinformation) or including (All) the most extreme of the deceptive sources --- those identified as spreading disinformation.} 
		\label{tab:breakdown}
 
	\end{table}
	
\subsection{Methodology}
We use the linguistically-infused neural network model from Figure~\ref{fig:model_architecture} to label the reaction type of each tweet or comment. Using these labels, we examine how often response types occur when users react to each type of news source. For clarity, we report the five most frequently occurring reaction types (expressed in at least 5\% of reactions within each source type) and compare the distributions of reaction types for each type of news source. 

To examine whether users react to content from trusted sources differently than from deceptive sources, we measure the reaction delay, which we define as the time elapsed between the moment the link or content was posted/tweeted and the moment that the reaction comment or tweet occurred. We report the cumulative distribution functions (CDFs) for each source type and use Mann Whitney U (MWU) tests to compare whether users respond with a given reaction type with significantly different delays to news sources of different levels of credibility.

\begin{figure}[t!]
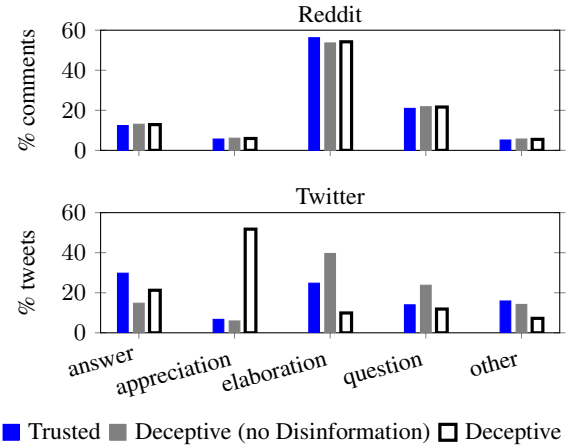

	\centering 
	\small    
	\small
%Reddit 


    \vspace{-2.1in}%-2.2in}
	\caption{Distributions of five most frequently occurring  reaction types within comments on Reddit and tweets on Twitter for each news source type (MWU $p<0.01$).}% (RQ1)} 
	\label{fig:coarse_discourse_label_distributions}
	\vspace{-0.3cm}
\end{figure}

\begin{figure}[t!]
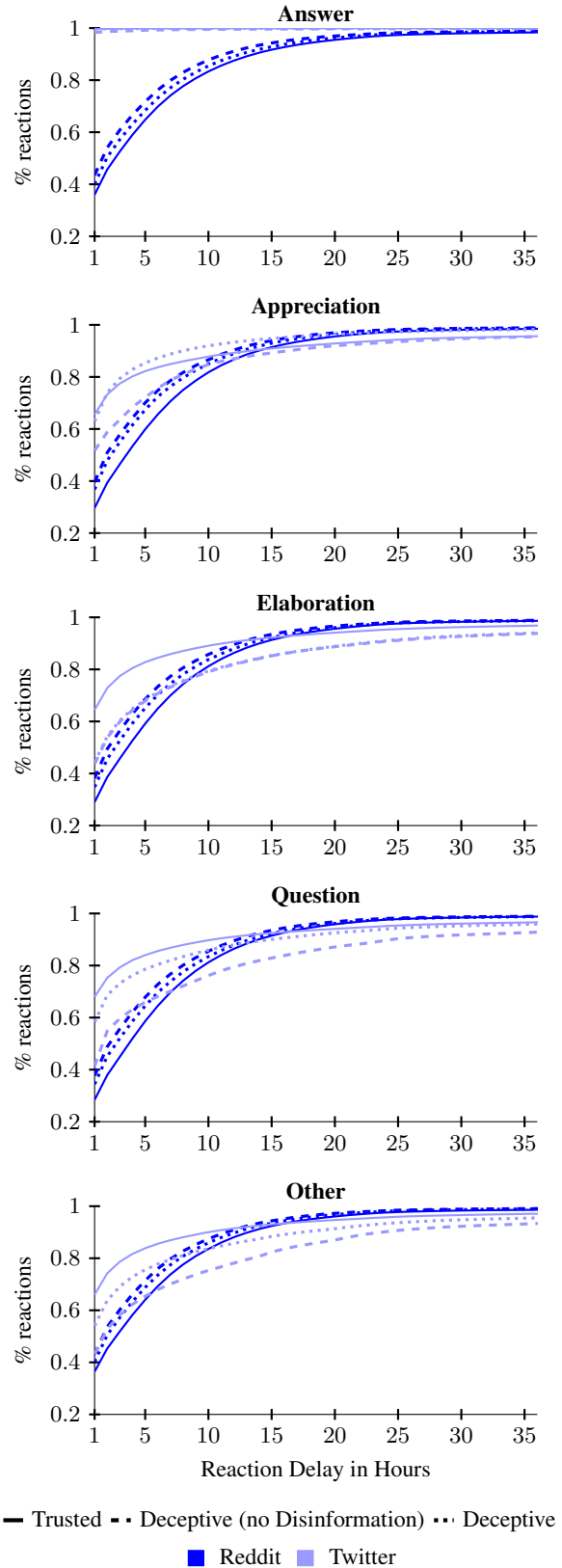

	\centering   
	\small 
	
	\small
%ANSWER
%

	\vspace{-2.1in}
	\caption{CDF plots of the volumes of reactions by reaction delays for the frequently occurring reactions (\ie, reactions that occur in at least 5\% of comments) for each source-type, using a step size of one hour. The CDF for Elaboration-reactions to Deceptive (no disinformation) Twitter news sources is occluded by the CDF for Deceptive Twitter news sources. This figure is best viewed in color.}% (RQ2)}       
	\label{fig:diffusion_delays_cdfs}
    \vspace{-.15in}
\end{figure}

\subsection{Results and Discussion}

For both Twitter and Reddit datasets, we found that the primary reaction types were answer, appreciation, elaboration, question, or ``other'' (no label was predicted). Figure~\ref{fig:coarse_discourse_label_distributions} illustrates the distribution of reaction types among Reddit comments (top plot) or tweets (bottom plot) responding to each type of source, as a percentage of all comments/tweets reacting to sources of the given type (\ie trusted, all deceptive, and deceptive excluding disinformation sources).

For Twitter, we report clear differences in user reactions to trusted vs. deceptive sources. {\it Deceptive  (including disinformation) sources have a much higher rate of appreciation reactions and a lower rate of elaboration responses, compared to trusted news sources}. Differences are still significant ($p<0.01$) but the trends reverse if we do not include disinformation sources. We also see an increase in the rate of question-reactions compared to trusted news sources if we exclude disinformation sources.  

For Reddit, there appears to be a very similar distribution across reaction types for trusted and deceptive sources. However, MWU tests still found that the differences between trusted and deceptive news sources were statistically significant ($p<0.01$) --- regardless of whether we include or exclude disinformation sources. 
{\it Posts that link to deceptive sources have higher rates of question, appreciation, and answering reactions, while posts that link to trusted sources have higher rates of elaboration, agreement, and disagreement.}

Next, we compared the speed with which users reacted to posts of sources of varying credibility. Our original hypothesis was that users react to posts of trusted sources faster than posts of deceptive sources. The CDFs for each source type and platform (solid and dashed lines represent Reddit and Twitter respectively) are  shown in Figure~\ref{fig:diffusion_delays_cdfs}. {\it We observe that the lifetime of direct reactions to news sources on Twitter is often more extended than for sources on Reddit. One exception is answer reactions which almost always occur within the first hour after the Twitter new source originally posted the tweet being answered.} This may be due to the different ways that users consume content on the two platforms. Users follow accounts on Twitter, whereas on Reddit users ``follow'' topics through their subscriptions to various subreddits. Users can view the news feeds of individual sources on Twitter and view all of the sources' posts. Reddit, on the other hand, is not designed to highlight individual users or news sources; instead new posts (regardless of the source) are viewed based on their hotness score within each subreddit.

In addition, we observe that {\it reactions to posts linked to trusted sources are less heavily concentrated within the first 12 to 15 hours of the post's lifetime on  Reddit. The opposite is found on Twitter. Twitter sources may have a larger range of reaction delays, but they are also more heavily concentrated in the lower end of that range ($p<0.01$)}.

\section{Related Work} 

As we noted above, most studies that examine misinformation spread focus on individual events such as natural disasters~\cite{takahashi2015communicating}, political elections~\cite{ferrara2017disinformation}, or crises~\cite{starbird2014rumors} and examine the response to the event on social media. 
A recent study by Vosoughi \etal \shortcite{vosoughi2018spread} found that news stories that were fact-checked and found to be false spread faster and to more people than news items found to be true.  In contrast, our methodology considers immediate reactions to news \textit{sources} of varying credibility, so we can determine whether certain reactions or reactions to trusted or deceptive news sources evoke more or faster responses from social media users.

\section{Conclusion}

In the current work, we have presented a content-based model that classifies user reactions into one of nine types, such as answer, elaboration, and question, etc., and a large-scale analysis of Twitter posts and Reddit comments in response to content from news sources of varying credibility.  

Our analysis of user reactions to trusted and deceptive sources on Twitter and Reddit shows significant differences in the distribution of reaction types for trusted versus deceptive news. However, due to differences in the user interface, algorithmic design, or user-base, we find that Twitter users react to trusted and deceptive sources very differently than Reddit users. For instance, Twitter users questioned disinformation sources less often and more slowly than they did trusted news sources; Twitter users also expressed appreciation towards disinformation sources more often and faster than towards trusted sources. Results from Reddit show similar, but far less pronounced, reaction results. 

Future work may focus on analysis of reaction behavior from automated (\ie 'bot'), individual, or organization accounts; on additional social media platforms and languages; or between more fine-grained categories of news source credibility.

\section*{Acknowledgments}
The research described in this paper is based on Twitter and Reddit data collected by the University of Notre Dame using public APIs. 
The research was supported by the
Laboratory Directed Research and Development Program at Pacific Northwest National Laboratory, a multiprogram national laboratory operated by Battelle for the U.S. Department of Energy. This research is also supported by the Defense Advanced Research Projects Agency (DARPA), contract W911NF-17-C-0094. The U.S. Government is authorized to reproduce and distribute reprints for Governmental purposes notwithstanding any copyright annotation thereon. The views and conclusions contained herein are those of the authors and should not be interpreted as necessarily representing the official policies or endorsements, either expressed or implied, of DARPA or the U.S. Government.

\bibliography{acl_deception_evolution_bib}
\bibliographystyle{acl_natbib}

\end{document}